\documentclass[runningheads]{llncs}
\usepackage{graphicx}

\usepackage{tikz}
\usepackage{comment}
\usepackage{amsmath,amssymb} 
\usepackage{color}
\usepackage{orcidlink}
\usepackage{graphicx}
\usepackage{amsmath}
\usepackage{amssymb}
\usepackage{booktabs}

\usepackage{algorithm}
\usepackage{algorithmic}

\usepackage{amssymb}
\usepackage{amsmath}
\usepackage{multirow}
\usepackage{newfloat}
\usepackage{listings}
\usepackage{color}
\usepackage[accsupp]{axessibility}  

\begin{document}
\pagestyle{headings}
\mainmatter
\def\ECCVSubNumber{100}  

\title{Prior-Guided Adversarial Initialization \\ 
for Fast Adversarial Training} 

\titlerunning{Prior-Guided Adversarial Initialization \\ 
for Fast Adversarial Training}
%
\author{Xiaojun Jia$^{1,2,\dagger}$, Yong Zhang$^{3,}$\thanks{  Correspondence to: Yong Zhang (\href{wubaoyuan@cuhk.edu.cn}{zhangyong201303@gmail.com}) and Xiaochun Cao (\href{mailto:caoxiaochun@mail.sysu.edu.cn}{caoxiaochun@mail.sysu.edu.cn}).$\dagger$ Work done in an internship at Tencent AI Lab.} Xingxing Wei$^{4}$, Baoyuan Wu$^{5}$, Ke Ma$^{6}$, \\ Jue Wang$^{3}$, Xiaochun Cao$^{1,7,\star}$\\}
%
%
\institute{$^{1}$SKLOIS, Institute of Information Engineering, CAS, Beijing, China\\
$^{2}$School of Cyber Security, University of Chinese Academy of Sciences, Beijing, China\\
$^{3}$ Tencent, AI Lab, Shenzhen, China \\
$^{4}$ Institute of Artificial Intelligence, Beihang University, Beijing, China. \\
$^{5}$School of Data Science, Secure Computing Lab of Big Data, Shenzhen Research Institute of Big Data, The Chinese University of Hong Kong, Shenzhen, China\\
$^{6}$ School of Computer Science and Technology, UCAS, Beijing, China \\
$^{7}$ School of Cyber Science and Technology, Shenzhen Campus, Sun Yat-sen University, Shenzhen 518107, China \\
{\tt\small jiaxiaojun@iie.ac.cn; zhangyong201303@gmail.com; xxwei@buaa.edu.cn;} \\ {\tt\small wubaoyuan@cuhk.edu.cn; make@ucas.ac.cn; arphid@gmail.com; caoxiaochun@mail.sysu.edu.cn}}
\authorrunning{Xiaojun Jia et al.}

\maketitle

\begin{abstract}
Fast adversarial training (FAT) effectively improves the efficiency of standard adversarial training (SAT).  
However, initial FAT encounters catastrophic overfitting, \textit{i.e.,} the robust accuracy against adversarial
attacks suddenly and dramatically decreases. 
Though several FAT variants spare no effort to prevent overfitting, they sacrifice much calculation cost. 
In this paper, we explore the difference between the training processes of SAT and FAT and observe that the attack success rate of adversarial examples (AEs) of FAT gets worse gradually in the late training stage, resulting in overfitting. 
The AEs are generated by the fast gradient sign method (FGSM)  with a zero or random initialization.   
Based on the observation, we propose a prior-guided FGSM initialization method to avoid overfitting after investigating several initialization strategies, improving the quality of the AEs during the whole training process. 
The initialization is formed by leveraging historically generated AEs without additional calculation cost. 
We further provide a theoretical analysis for the proposed initialization method. 
We also propose a simple yet effective regularizer based on the prior-guided initialization, \textit{i.e.,} the currently generated perturbation should not deviate too much from the prior-guided initialization. 
The regularizer adopts both historical and current adversarial perturbations to guide the model learning.
Evaluations on four datasets demonstrate that the proposed method can prevent catastrophic overfitting and outperform state-of-the-art FAT methods. The code is released at
\textcolor{red}{{\url{https://github.com/jiaxiaojunQAQ/FGSM-PGI}}}.

\end{abstract}
\section{Introduction}
Deep neural networks (DNNs) \cite{lecun2015deep,bai2021improving,zou2019reinforced,gu2021capsule,wei2018transferable,wang2021enhancing,jia2019comdefend,liang2020efficient,liang2022parallel,li2022semi} are vulnerable to adversarial examples (AEs) \cite{szegedy2013intriguing,chen2018robust,finlayson2019adversarial,duan2020adversarial,lin2019nesterov,xie2019improving,jia2020adv,gu2021effective,duan2021adversarial,bai2020targeted,bai2020improving} generated by adding imperceptible perturbations to benign images.
Standard adversarial training (SAT) \cite{madry2017towards,wang2019improving,jia2022adversarial} is one of the most efficient defense methods against AEs. 
It adopts projected gradient descent (PGD) ~\cite{madry2017towards}, a multi-step attack method, to generate AEs for training.
But, SAT requires much time to calculate gradients of the network's input multiple times.

\begin{figure}[t]

        \centering
        \includegraphics[width=0.82\linewidth]{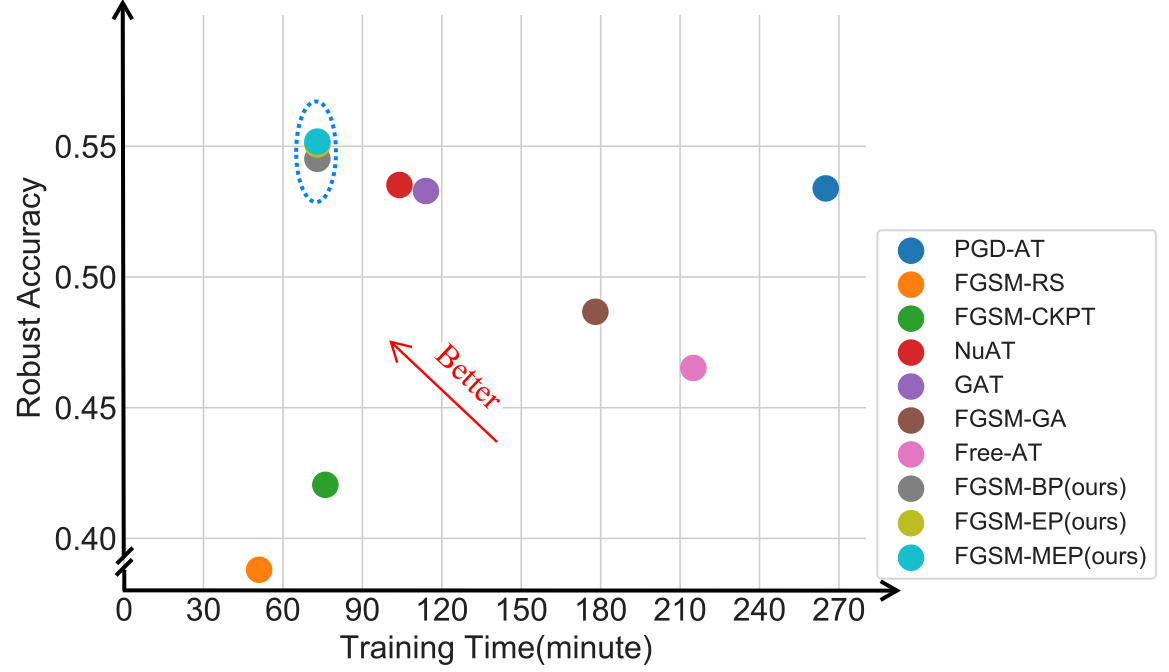}
        \label{fig:time_acc}

  \caption{ PGD-10 robust accuracy and training time of different FAT methods with ResNet18 as the backbone on the CIFAR-10 dateset. $x$-axis denotes training time (lower is more efficient) and $y$-axis denotes PGD-10 accuracy (higher is more robust).  }
\label{fig:time_acc}

\end{figure}
\par To improve efficiency, fast adversarial training (FAT) methods \cite{goodfellow2014explaining,park2021reliably,zhang2019you,jia2022boosting} have been proposed. Goodfellow~\emph{et al.} first  \cite{goodfellow2014explaining} adopt FGSM to generate AEs for training, \textit{i.e.,} FGSM-AT, but it encounters the catastrophic overfitting. 
Wong \emph{et al.} \cite{wong2020fast} propose to combine FGSM-AT with random initialization along with early stopping to overcome the overfitting. 
Andriushchenko \emph{et al.} \cite{andriushchenko2020understanding}
propose a method from the view of regularization to enhance the quality of AEs, \textit{i.e.,} GradAlign. 
Kim \emph{et al.} \cite{kim2020understanding} propose a simple method to determine an appropriate step size of FGSM-AT to generate stronger AEs, \textit{i.e.,} FGSM-CKPT. 
Sriramanan \emph{et al.} \cite{sriramanan2020guided} introduce a relaxation term in the training loss to improve the quality of the generated AEs, \textit{i.e.,} GAT. 
Then, Sriramanan \emph{et al.}~\cite{sriramanan2021towards} propose a Nuclear-Norm regularizer to enhance
the optimization of the AE generation and the model training. 
These methods not only prevent
catastrophic overfitting but also achieve advanced defense performance by enhancing the quality of AEs.
However, compared with FGSM-RS, these advanced methods require high additional calculation costs to generate stronger AEs.

\begin{figure}[t]
\centering
 \includegraphics[width=0.95\linewidth]{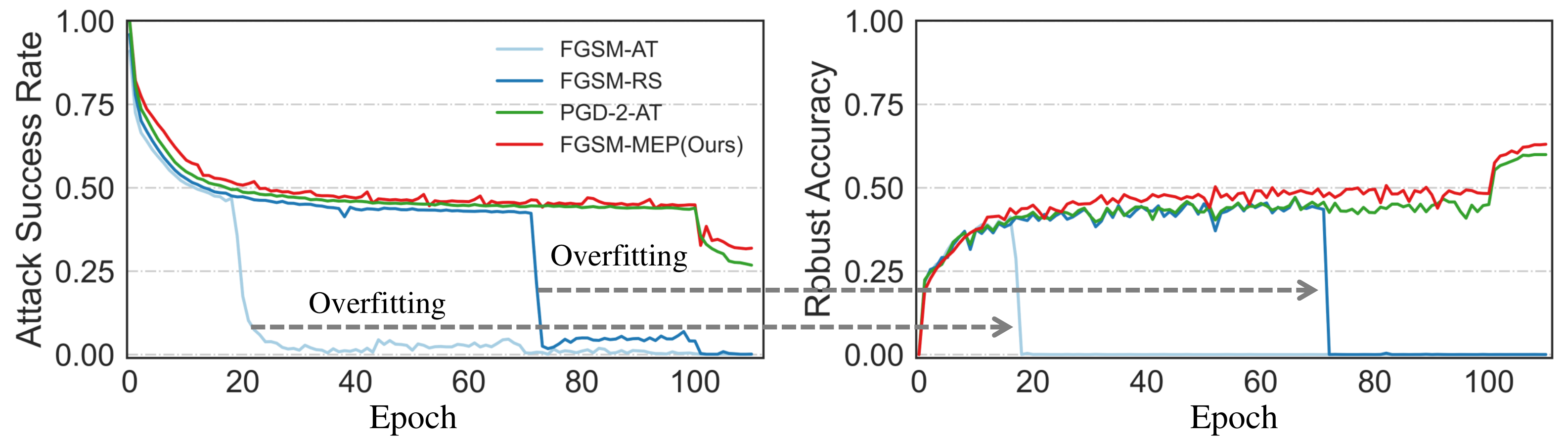}

  \caption{
  The difference between the training processes of FAT and SAT methods on the CIFAR-10 dataset.
  Left: the attack success rate of generated AEs. 
  Right: the PGD-10 robust accuracy of the target model. 
  FGSM-AT and FGSM-RS encounter catastrophic overfitting after training a few epochs. 
  But PGD-2-AT and our method can prevent the overfitting. 
  }
\label{fig:difference}
\end{figure}

As AEs are critical for adversarial training, we study the AEs of both FAT and SAT during their training processes to explore the reason for catastrophic overfitting. 
Surprisingly, we observe a distinct difference between the attack success rates of their AEs in the late training stage. 
The attack success rate of FAT drops sharply after a few training epochs while the robust accuracy decreases to 0 (see FGSM-AT and FGSM-RS in Fig.~\ref{fig:difference}).
This phenomenon indicates that  overfitting happens when the quality of AEs becomes worse. 
However, the overfitting does not exist in the two-step PGD (PGD-2-AT) that can be treated as FGSM-AT with an adversarial initialization. It means adversarial initialization could be a solution to overfitting.  
Based on the observation, we raise a question \textit{``can we obtain an adversarial initialization for FGSM-AT to maintain the high quality of AEs for avoiding catastrophic overfitting without additional calculation cost?"}
We investigate several initialization strategies and propose a prior-guided initialization by leveraging historically generated AEs, dubbed FGSM-PGI. 
Specifically, we first use the buffered perturbations of AEs from the previous batch and the previous epoch as the initialization for FGSM, respectively, called FGSM-BP and FGSM-EP. The two strategies are demonstrated to be effective. 
 To exploit complete prior information, 
 we then propose to leverage the buffered gradients from all previous epochs via a momentum mechanism as an additional prior, dubbed FGSM-MEP, which works the best. 
Furthermore, we provide a theoretical analysis for the proposed initialization method. 
Moreover, we also propose a simple yet effective regularizer based on the above prior-guided initialization to guide model learning for better robustness. 
The current perturbation can be generated via FGSM using the prior-guided initialization. 
The regularizer prevents the current perturbation
deviating too much from the prior-guided initialization, which is implemented by using a squared $L_2$ distance between the predictions of adversarial examples generated based on the current perturbation and the prior-guided initialization.

The proposed method not only prevents catastrophic overfitting but also improves the adversarial robustness against adversarial attacks.  
Fig.~\ref{fig:time_acc} illustrates the robustness and training efficiency of different FAT methods.
Our main contributions are in three aspects:
\textbf{1)} We propose a prior-guided adversarial initialization to prevent overfitting after investigating several initialization strategies. 
\textbf{2)} We also propose a regularizer to guide the model learning for better robustness by considering both the currently generated perturbation and the prior-guided initialization.
\textbf{3)} Extensive experiments on four datasets demonstrate that the proposed method can outperform state-of-the-art FAT methods in terms of both efficiency and robustness. 

\section{Related Work}
\subsection{Adversarial Training}
Adversarial training (AT) variants \cite{zhang2019theoretically,rice2020overfitting,DBLP:conf/nips/WuX020} are effective in defending against AEs. 
Madry \emph{et al.}~\cite{madry2017towards} first formulate AT as a minimax optimization problem formulated as follows:
\begin{equation}
\min _{\mathbf{w}} \mathbb{E}_{(\mathbf{x}, \mathbf{y}) \sim \mathcal{D}}[\max _{{\boldsymbol{\delta}} \in \Omega}\mathcal{L}(f(\mathbf{x}+{\boldsymbol{\delta}} ; \mathbf{w}), \mathbf{y})],
\label{Eq:minimax}
\end{equation}
where $f(\cdot ; \mathbf{w})$ represents a deep neural network with parameters $\mathbf{w}$, $\mathcal{D}$ represents a joint data distribution of the benign image $\mathbf{x}$ and the GT one-hot label $\mathbf{y}$.
$\mathcal{L}(f(\mathbf{x} ; \mathbf{w}), \mathbf{y})$ represents the classification loss function.  $\mathbf{\boldsymbol{\delta}}$ represents the adversarial perturbation, and $\Omega$ represents a threat bound which can be defined as $ \Omega = \{\mathbf{\boldsymbol{\delta}}:\|\mathbf{\boldsymbol{\delta}}\| \leq \epsilon\}$ with the maximum perturbation strength $\epsilon$. 
Multi-step AT methods generate AEs by a multi-step adversarial attack method, \textit{i.e.}, projected gradient ascent (PGD) \cite{madry2017towards}. It can be defined as:
\begin{equation}
\mathbf{\boldsymbol{\delta}}_{t+1}=\Pi_{[-\epsilon, \epsilon]}\left[\mathbf{\boldsymbol{\delta}}_{t}+\alpha \cdot \textrm{sign} \left(\nabla_{\mathbf{x}} \mathcal{L}\left(f\left(\mathbf{x}+\mathbf{\boldsymbol{\delta}}_{t} ; \mathbf{w}\right), \mathbf{y}\right)\right)\right],
\end{equation}
where $\mathbf{\boldsymbol{\delta}}_{t+1}$ represents the perturbation at the $t+1$-th iteration, $\alpha$ represents the attack step size, and $\Pi_{[-\epsilon, \epsilon]}$ is the projection that maps the input to the range $[-\epsilon, \epsilon]$. This is a prime PGD-based AT framework proposed in \cite{madry2017towards}. 
Following this work, 
many advanced variants \cite{DBLP:conf/icml/WangM0YZG19,DBLP:conf/iclr/0001ZY0MG20,DBLP:conf/iclr/BaiZJXM021,kannan2018adversarial,pang2020boosting,roth2019adversarial} have been proposed from different perspectives to improve model robustness. 
An early stopping version \cite{rice2020overfitting} of PGD-based AT stands out amongst them. 

\subsection{Fast Adversarial Training}
\par Although multi-step AT methods can achieve good robustness, they require lots of calculation costs to generate AEs for training. 
Fast adversarial training variants that generate AEs by the one-step fast gradient sign method (FGSM) \cite{goodfellow2014explaining} are proposed to improve the efficiency, which can be dubbed FGSM-based AT methods.
The perturbation of FGSM-AT \cite{goodfellow2014explaining} is defined as:
\begin{equation}
\mathbf{\boldsymbol{\delta}}=\epsilon \cdot \textrm{sign}\left(\nabla_{\mathbf{x}} \mathcal{L}(f(\mathbf{x} ; \mathbf{w}), \mathbf{y})\right),
\label{Eq:FGSM_AT}
\end{equation}
where $\epsilon$ is the maximal perturbation strength. 
Though FGSM-based AT accelerates the training speed of AT, it encounters catastrophic overfitting. 
Specifically, after a period of training, the trained model suddenly cannot defend against the AEs generated by PGD. 
Wong \emph{et al.}~\cite{wong2020fast} propose to conduct FGSM-based AT with a random initialization. 
Its perturbation can be defined as: 
\begin{equation}
\mathbf{\boldsymbol{\delta}}=\Pi_{[-\epsilon, \epsilon]}\left[\mathbf{\boldsymbol{\boldsymbol{\eta}}}+\alpha \cdot \textrm{sign}\left(\nabla_{\mathbf{x}} \mathcal{L}(f(\mathbf{x}+\mathbf{\boldsymbol{\boldsymbol{\eta}}} ; \mathbf{w}), \mathbf{y})\right)\right],
\label{eq:FGSM_RS}
\end{equation}
where  $\boldsymbol{\boldsymbol{\eta}} \in \mathbf{U}(-\epsilon, \epsilon)$ represents the random initialization, $\mathbf{U}$ is a uniform distribution, and $\alpha$ is set to $1.25\epsilon$. After that, a series of FAT methods enhances the quality of AEs to improve the defense performance. In detail, 
Andriushchenko \emph{et al.} \cite{andriushchenko2020understanding} demonstrate that using the random initialization cannot prevent catastrophic overfitting. 
They propose a regularizer to generate stronger AEs, \textit{i.e.,} FGSM-GA. 
Kim \emph{et al.}~\cite{kim2020understanding} reveal that FGSM-RS adopts AEs with the maximum perturbation instead of ones in the adversarial direction, resulting in the overfitting. They then propose a method to determine an appropriate step size to generate powerful AEs to improve model robustness. 
Sriramanan \emph{et al.} \cite{sriramanan2020guided} enhance the quality of the generated AEs by using a relaxation term for the classification loss, \textit{i.e.,} GAT. Besides, Sriramanan \emph{et al.}~\cite{sriramanan2021towards} design a Nuclear-Norm regularizer to enhance
optimization of the AEs generation to improve the model robustness, \textit{i.e.,} NuAT. 
Though these methods can prevent overfitting and achieve advanced defense performance, they require much additional calculation cost to conduct AT.

\section{The Proposed Approach}
We first present our observations of revisiting catastrophic overfitting in Sec.~\ref{sec:Overfitting} and then explore several prior-guided adversarial initialization  strategies for FGSM-AT to prevent the overfitting in Sec.~\ref{sec:Initialization}, \textit{i.e.,} FGSM-PGI. We propose a simple yet effective regularization method for the FGSM-PGI to further improve model robustness in Sec.~\ref{sec:regularization}. Moreover, we also provide a theoretical analysis for the proposed initialization method in Sec.~\ref{sec:theoretical}.

\subsection{Revisting Catastrophic Overfitting}
\label{sec:Overfitting}
Catastrophic overfitting is a phenomenon that the robust accuracy of FAT drops sharply to 0\% in the late training stage (see the left figure of Fig.~\ref{fig:difference}), which was noticed by Wong \textit{et al.}~\cite{wong2020fast}. 
They then found random initialization of FGSM can help avoid the overfitting with training for limited epochs. 
However, if the training process of FGSM-RS goes on, catastrophic overfitting still appears, as revealed in ~\cite{andriushchenko2020understanding}, which indicates that random initialization must be conducted with early stopping. 
Hence, FGSM-RS does not solve the overfitting fundamentally.
FGSM-GA~\cite{andriushchenko2020understanding} proposes a method from the perspective of regularization in the loss to generate stronger AEs. FGSM-CKPT~\cite{kim2020understanding} proposes a method for adjusting the step size of FGSM to improve the quality of AEs in different training stages. 
GAT~\cite{sriramanan2020guided} adopts a relaxation term for the classification loss to generate stronger AEs during the AEs generation. NuAT~\cite{sriramanan2021towards} uses a Nuclear Norm regularizer to improve the optimization of the AEs generation.
These methods prevent catastrophic overfitting by improving the quality of AEs during the training.
However, FGSM-GA requires calculating gradients multiple times for regularization, FGSM-CKPT requires multiple times of forwarding propagation for step size selection, and GAT and NuAT require much more time to conduct regularization for generating powerful AEs, which significantly reduces efficiency. 

We reinvestigate catastrophic overfitting by comparing the intermediate results of the training processes of FAT and SAT on the CIFAR-10 dataset~\cite{krizhevsky2009learning}.  
For FAT, we study two methods, \textit{i.e.,} FGSM-AT and FGSM-RS. 
For SAT, we study PGD-AT~\cite{madry2017towards} with the iteration time of 2, \textit{i.e.,} PGD-2-AT.
As AEs are the key that distinguishes adversarial training from conventional training, we observe the quality of adversarial examples during the whole training process, \textit{i.e.,} whether adversarial examples can successfully attack the target model. 
The quality is evaluated by attack success rate (ASR). 
Fig.~\ref{fig:difference} illustrates the curves of the robust accuracy and the ASR of the above methods as well as ours. 
We also observe similar phenomenons on other benchmark datasets that are shown in the \textbf{supplementary material}.   

We summarize the observations as follows. First, it can be observed that the ASRs of FGSM-AT and FGSM-RS drop sharply at the 20-th and 74-th epoch, respectively, leading to the dramatic decreases of the robust accuracy. 
This indicates that the model robustness would collapse against adversarial attacks if the generated adversarial examples fail in attacking against the model during training.  
Note that the investigation from the perspective of the quality of adversarial example (\textit{i.e.,} ASR) is ignored by previous works. 
Second, FGSM-RS with random initialization delays the overfitting from the 20-th epoch to the 70-th epoch. 
Enhancing the initialization could alleviate the overfitting, but cannot essentially prevent it, which is also observed in ~\cite{andriushchenko2020understanding}. 
Third, surprisingly, PGD-2-AT does not suffer from the overfitting, which draws our attention. 
PGD-2-AT can be treated as FGSM-AT with an adversarial initialization of FGSM for adversarial example generation. 
Even after 110 training epochs, there are still a portion of high-quality adversarial examples that can fool the model. 
It indicates that adversarial initialization can improve the quality of adversarial examples. 
The initialization requires additional gradient calculation, which is not desirable.

\begin{table}[]

\centering
\caption{Comparisons of clean and robust accuracy (\%) and training time (minute) on the  CIFAR-10 dataset. Number in bold indicates the best.
}
\label{table:FGSM_MEP}
\scalebox{0.9}{

\begin{tabular}{@{}c|ccccccc|c@{}}
\toprule
Method                    &      & Clean          & PGD-10         & PGD-20         & PGD-50         & C\&W           & AA             & Time(min)            \\ \midrule
\multirow{2}{*}{FGSM-BP}  & Best & \textbf{83.15} & 54.59          & 53.55          & 53.2           & 50.24          & 47.47          & \multirow{2}{*}{73}  \\ \cmidrule(lr){2-8}
                          & Last & \textbf{83.09} & 54.52          & 53.5           & 53.33          & 50.12          & 47.17          &                      \\ \midrule
\multirow{2}{*}{FGSM-EP}  & Best & 82.75          & 54.8           & 53.62          & 53.27          & 49.86          & 47.94          & \multirow{2}{*}{73}  \\ \cmidrule(lr){2-8}
                          & Last & 81.27          & 55.07          & 54.04          & 53.63          & 50.12          & 46.83          &                      \\ \midrule
\multirow{2}{*}{FGSM-MEP} & Best & 81.72          & \textbf{55.18} & \textbf{54.36} & \textbf{54.17} & \textbf{50.75}          & \textbf{49.00} & \multirow{2}{*}{73}  \\ \cmidrule(lr){2-8}
                          & Last & 81.72          & \textbf{55.18} & \textbf{54.36} & \textbf{54.17} & \textbf{50.75}          & \textbf{49.00} &                      \\ \bottomrule
\end{tabular}
}
\end{table}

\subsection{Prior-Guided Adversarial Initialization}
\label{sec:Initialization}
The above three observations motivate us to dive into the question \textit{``how to obtain adversarial initialization without additional calculation cost"}. 
We come up with using the adversarial perturbations of historical AEs as the initialization for FGSM to conduct adversarial training, dubbed FGSM-PGI. 
Such perturbations serve as the prior knowledge that we can freely obtain without additional gradient calculation except for extra storage.  
We explore three strategies of exploiting historical perturbations, \textit{i.e.,} taking perturbations from the previous batch, from the previous epoch, and from the momentum of all epochs, \textit{i.e.,} FGSM-BP, FGSM-EP, and FGSM-MEP.

\noindent\textbf{Prior From the Previous Batch (FGSM-BP).}
We store the perturbations of AEs from the previous batch and regard them as the initialization of FGSM to generate AEs in the current batch. 
As the batch is randomly sampled, there is no correspondence between perturbations from the previous batch and samples from the current batch.  
Specifically, for a data point $\mathbf{x}$, we first add the perturbation from the previous batch onto it and then conduct FGSM based on the perturbed example. 
The adversarial perturbation can be defined as:
\begin{equation}
\label{eq:FSGM_BP}
\mathbf{\boldsymbol{\delta}}_{B_{t+1}}=\Pi_{[-\epsilon, \epsilon]}\left[\mathbf{\boldsymbol{\delta}}_{B_{t}}+\alpha \cdot \operatorname{sign}\left(\nabla_{\mathbf{x}} \mathcal{L}(f(\mathbf{x}+\mathbf{\boldsymbol{\delta}}_{B_{t}} ; \mathbf{w}), \mathbf{y})\right)\right],
\end{equation}
where $\mathbf{\boldsymbol{\delta}}_{B_{t+1}}$ is the adversarial perturbation in the $t+1$-th batch. 
Compared with FGSM-RS (see Eq.~\ref{eq:FGSM_RS}), we replace the random initialization with the adversarial perturbation from the previous batch.

\noindent\textbf{Prior From the Previous Epoch (FGSM-EP).}
We store the perturbations of all adversarial samples from the previous epoch and use them as initialization of FGSM to generate adversarial perturbations for the samples in the current epoch. 
Note that there is a correspondence between each perturbation from the previous epoch and the sample in the current epoch.  
 The adversarial perturbation is defined as:
\begin{equation}
    \label{eq:FSGM_EP}
    \mathbf{\boldsymbol{\delta}}_{E_{t+1}}=\Pi_{[-\epsilon, \epsilon]}\left[\mathbf{\boldsymbol{\delta}}_{E_{t}}+\alpha \cdot \operatorname{sign}\left(\nabla_{\mathbf{x}} \mathcal{L}(f(\mathbf{x}+\mathbf{\boldsymbol{\delta}}_{E_{t}} ; \mathbf{w}), \mathbf{y})\right)\right],
\end{equation}
where $\mathbf{\boldsymbol{\delta}}_{E_{t+1}}$ is the adversarial perturbation in the $t+1$-th epoch.  
Compared with FGSM-RS (see Eq.~\ref{eq:FGSM_RS}), we replace the random initialization with the adversarial perturbation from the previous epoch.

\vspace{1mm}
\noindent\textbf{Prior From the Momentum of All Previous Epochs (FGSM-MEP).}
To completely leverage historical adversarial perturbations during the whole training process, we propose to compute the momentum of one sample's gradients across all previous training epochs. 
Then, the gradient momentum is used as the initialization of FGSM for AE generation in the current epoch. 
There is also a correspondence between the gradient momentum and the sample in the current epoch. 
The adversarial perturbation can be defined as: 
\begin{align}
&\mathbf{g}_{c} =\operatorname{sign}\left(\nabla_{\mathbf{x}} \mathcal{L}(f(\mathbf{x}+\mathbf{\boldsymbol{\eta}}_{E_t} ; \mathbf{w}), \mathbf{y})\right), \\
&\mathbf{g}_{E_{t+1}} =\mu \cdot \mathbf{g}_{E_{t}} + \mathbf{g}_{c}, \\
&\mathbf{\boldsymbol{\delta}}_{E_{t+1}}=\Pi_{[-\epsilon, \epsilon]}\left[\mathbf{\boldsymbol{\eta}}_{E_{t}}+\alpha \cdot \mathbf{g}_{c} \right], \label{eq:FSGM_MEP_1}\\
&\mathbf{\boldsymbol{\eta}}_{E_{t+1}} =\Pi_{[-\epsilon, \epsilon]}\left[\mathbf{\boldsymbol{\eta}}_{E_{t}}+\alpha \cdot \operatorname{sign}(\mathbf{g}_{E_{t+1}}) \right]. \label{eq:FSGM_MEP_2}
\end{align}
Similar to FGSM-AT (Eq.~\ref{Eq:FGSM_AT}), $\mathbf{g}_c$ is regarded as the signed gradient and $\mathbf{g}_{E_{t+1}}$ is the signed gradient momentum in the $t+1$-th epoch.   
$\mu$ represents the decay factor. 
${\boldsymbol{\delta}}_{E_{t+1}}$ is the adversarial perturbation, similar to that of FGSM-EP. 
${\boldsymbol{\eta}}_{E_{t+1}}$ is the projected perturbation, which is used as adversarial initialization in the next epoch.  
Compared with FGSM-EP, FGSM-MEP exploits the gradient momentum involving information of all previous epochs as initialization, instead of using only information in the previous epoch.

\subsection{Prior-guided Initialization based Regularization }
\label{sec:regularization}
We propose an effective regularization method based on the prior-guided initialization to improve model robustness.
Specifically, given the prior-guided initialization, we can generate a current perturbation via FGSM with the initialization. 
Both the current perturbation and the initialization can be used to create adversarial examples. 
Forcing the two types of adversarial examples to have similar predictions could help improve the smoothness of function. 
The proposed regularization term can be added into the training loss to update the model
parameters $\mathbf{w}_{t+1}$, as follows:
\begin{equation}
\mathbf{w}_{t+1}=\arg\min_{\mathbf{w}} [\mathcal{L}(f(\mathbf{x}+\boldsymbol{\delta}_{adv}; \mathbf{w}), \mathbf{y}) +
\lambda \cdot\left\|f(\mathbf{x}+\boldsymbol{\delta}_{adv} ; \mathbf{w})-{f}(\mathbf{x}+\boldsymbol{\delta}_{pgi}; \mathbf{w})\right\|_{2}^{2}],
\label{eq:loss}
\end{equation} 
where $\boldsymbol{\delta}_{pgi}$ represents the prior-guided initialization generated by one of the above three initialization methods. 
$\boldsymbol{\delta}_{adv}$ represents the current adversarial perturbation generated via FGSM using $\boldsymbol{\delta}_{pgi}$ as initialization.  
$\lambda$ is a trade-off hyper-parameter. 
The first term is the cross-entropy loss on AEs generated using the current perturbation.   
The second term represents the regularization, \textit{i.e.,} the squared $L_{2}$ distance between the predictions of the two types of adversarial examples.  
This term makes the learned model not only robust to currently generated AEs but also historically generated AEs. 
In this way, the proposed regularization term explicitly enforces the function smoothness around samples to improve model robustness. 

\par Based on the proposed prior-guided adversarial initialization and the regularization, we can establish our FAT framework. We evaluate the three prior-guided adversarial initialization approaches on CIFAR-10 (see results in Table~\ref{table:FGSM_MEP}) and find that they all can prevent catastrophic overfitting and achieve advanced model robustness against adversarial attacks.  
FGSM-MEP works the best in terms of robust accuracy under all attack scenarios.
\textbf{The FGSM-EP, FGSM-BP and FGSM-MEP algorithms are presented in the supplementary material}.

\subsection{Theoretical Analysis}
\label{sec:theoretical}
\begin{proposition}
    Let $\boldsymbol{\delta}_{pgi}$ be the prior-guided adversarial initialization in \textbf{FGSM-BP}, \textbf{FSGM-EP} or \textbf{FSGM-MEP}, $\boldsymbol{\hat{\delta}}_{adv}$ represents the current adversarial perturbation generated via FGSM using $\boldsymbol{\delta}_{pgi}$ as initialization, and $\alpha$ be the step size of \eqref{eq:FSGM_BP}, \eqref{eq:FSGM_EP}, \eqref{eq:FSGM_MEP_1} and \eqref{eq:FSGM_MEP_2}. If $\boldsymbol{\Omega}$ is a bounded set like
    \begin{equation}
        \boldsymbol{\Omega} = \big\{\boldsymbol{\hat{\delta}}_{adv}\ :\ \|\boldsymbol{\hat{\delta}}_{adv}-\boldsymbol{\delta}_{pgi}\|^2_2\leq{\epsilon}^{2}\big\},
    \end{equation}
    and the step size $\alpha$ satisfies $\alpha\leq\epsilon$, it holds that
    \begin{equation}
        \label{eq:propo_1}
        \begin{aligned}
            & \mathbb{E}_{\boldsymbol{\hat{\delta}}_{adv}\sim\boldsymbol{\Omega}}\big[\|\boldsymbol{\hat{\delta}}_{adv}\|_2\big]&\leq&\ \ \sqrt{\ \mathbb{E}_{\boldsymbol{\hat{\delta}}_{adv}\sim\boldsymbol{\Omega}}\big[\|\boldsymbol{\hat{\delta}}_{adv}\|^2_2\big]}\\
            & &\leq&\ \ \sqrt{\frac{1}{d}}\cdot\epsilon,
        \end{aligned}
    \end{equation}
    where $\boldsymbol{\hat{\delta}}_{adv}$ is the adversarial perturbation generated by \textbf{FGSM-BP}, \textbf{FSGM-EP} or \textbf{FSGM-MEP}, and $d$ is the dimension of the feature space. 
\end{proposition}
The proof is deferred to the \textbf{supplementary material}. The upper bound of the proposed method is $\sqrt{\frac{1}{d}}\cdot\epsilon$ which is less than the bound $\sqrt{\frac{d}{3}}\cdot\epsilon$ of FGSM-RS  provided in \cite{andriushchenko2020understanding}. Due to the norm of perturbation (gradient) can be treated as the convergence criteria for the non-convex optimization problem, the smaller expectation represents that the proposed prior-guided adversarial initialization will be converged to a local optimal faster than the random initialization with the same number of iterations.

\section{Experiments}
\label{Experiments}

\begin{table*}[t]
\centering
\caption{Comparisons of clean and robust accuracy (\%) and training time (minute) using ResNet18 on the CIFAR-10 dataset. Number in bold indicates the best. 
}
\label{table:cifar10_ResNet}

 \scalebox{0.85}{
\begin{tabular}{@{}c|ccccccc|c@{}}
\toprule
Method                         &      & Clean & PGD-10         & PGD-20         & PGD-50         & C\&W           & AA             & Time(min)            \\ \midrule
\multirow{2}{*}{PGD-AT~\cite{rice2020overfitting}}          & Best & 82.32 & 53.76          & 52.83          & 52.6           & 51.08          & 48.68          & \multirow{2}{*}{265} \\ \cmidrule(lr){2-8}
                                 & Last & 82.65 & 53.39          & 52.52          & 52.27          & 51.28          & 48.93          &                      \\ \midrule \midrule
\multirow{2}{*}{FGSM-RS~\cite{wong2020fast}}         & Best & 73.81 & 42.31          & 41.55          & 41.26          & 39.84          & 37.07          & \multirow{2}{*}{51}  \\ \cmidrule(lr){2-8}
                                 & Last & 83.82 & 00.09          & 00.04          & 00.02          & 0.00           & 0.00           &                      \\ \midrule
\multirow{2}{*}{FGSM-CKPT~\cite{kim2020understanding}}       & Best & \textbf{90.29} & 41.96          & 39.84          & 39.15          & 41.13          & 37.15          & \multirow{2}{*}{76}  \\ \cmidrule(lr){2-8}
                                 & Last & \textbf{90.29} & 41.96          & 39.84          & 39.15          & 41.13          & 37.15          &                      \\ \midrule
\multirow{2}{*}{NuAT~\cite{sriramanan2021towards}}            & Best & 81.58 & 53.96          & 52.9           & 52.61          & \textbf{51.3}  & \textbf{49.09} & \multirow{2}{*}{104}  \\ \cmidrule(lr){2-8}
                                 & Last & 81.38 & 53.52          & 52.65          & 52.48          & 50.63          & 48.70          &                      \\ \midrule
\multirow{2}{*}{GAT~\cite{sriramanan2020guided}}             & Best & 79.79 & 54.18          & 53.55          & 53.42          & 49.04          & 47.53          & \multirow{2}{*}{114} \\ \cmidrule(lr){2-8}
                                 & Last & 80.41 & 53.29          & 52.06          & 51.76          & 49.07          & 46.56          &                      \\ \midrule
\multirow{2}{*}{FGSM-GA~\cite{andriushchenko2020understanding}}         & Best & 83.96 & 49.23          & 47.57          & 46.89          & 47.46          & 43.45          & \multirow{2}{*}{178} \\ \cmidrule(lr){2-8}
                                 & Last & 84.43 & 48.67          & 46.66          & 46.08          & 46.75          & 42.63          &                      \\ \midrule
\multirow{2}{*}{Free-AT(m=8)~\cite{shafahi2019adversarial}}    & Best & 80.38 & 47.1           & 45.85          & 45.62          & 44.42          & 42.17          & \multirow{2}{*}{215} \\ \cmidrule(lr){2-8}
                                 & Last & 80.75 & 45.82          & 44.82          & 44.48          & 43.73          & 41.17          &                      \\ \midrule
\multirow{2}{*}{FGSM-BP (ours)}  & Best & 83.15 & 54.59          & 53.55          & 53.2           & 50.24          & 47.47          & \multirow{2}{*}{73}  \\ \cmidrule(lr){2-8}
                          & Last & 83.09 & 54.52          & 53.5           & 53.33          & 50.12          & 47.17          &                      \\ \midrule
\multirow{2}{*}{FGSM-EP (ours)}  & Best & 82.75          & 54.8           & 53.62          & 53.27          & 49.86          & 47.94          & \multirow{2}{*}{73}  \\ \cmidrule(lr){2-8}
                          & Last & 81.27          & 55.07          & 54.04          & 53.63          & 50.12          & 46.83          &                      \\ \midrule                      
\multirow{2}{*}{FGSM-MEP (ours)} & Best & 81.72 & \textbf{55.18} & \textbf{54.36} & \textbf{54.17} & 50.75          & 49.00          & \multirow{2}{*}{73}  \\ \cmidrule(lr){2-8}
                                 & Last & 81.72 & \textbf{55.18} & \textbf{54.36} & \textbf{54.17} & \textbf{50.75} & \textbf{49.00} &                      \\ \bottomrule
\end{tabular}
}

\end{table*}
\subsection{Experimental Setting}
\noindent \textbf{Dataset Settings.} To evaluate the effectiveness of the proposed FGSM-MEP, extensive experiments are conducted on four benchmark datasets that are the most widely used to evaluate adversarial robustness, \textit{i.e.}, CIFAR-10~\cite{krizhevsky2009learning}, CIFAR-100~\cite{krizhevsky2009learning}, Tiny ImageNet~\cite{deng2009imagenet}, and ImageNet~\cite{deng2009imagenet}. 
Following the commonly used settings in the adversarial training, we adopt ResNet18~\cite{he2016deep} and WideResNet34-10~\cite{DBLP:conf/bmvc/ZagoruykoK16} as the backbone on CIFAR-10 and CIFAR-100,  PreActResNet18~\cite{he2016identity} on Tiny ImageNet, and ResNet50~\cite{he2016deep} on ImageNet. 
We adopt the SGD optimizer \cite{DBLP:journals/nn/Qian99} with a learning rate of 0.1,  the weight decay of 5e-4, and the momentum of 0.9.
As for CIFAR-10, CIFAR-100, and Tiny ImageNet, following the settings of \cite{rice2020overfitting,pang2020bag},
the total epoch number is set to 110. The learning rate decays with a factor of 0.1 at the 100th and 105th epoch. 
As for ImageNet, following the settings of \cite{shafahi2019adversarial,wong2020fast}, the total epoch number is set to 90. 
The learning rate decays with a factor of 0.1 at the 30th and 60th epoch.
Experiments on ImageNet are conducted with 8 Tesla V100 and other experiments are conducted with a single Tesla V100. 
For our hyper-parameters, the decay factor $\mu$ is set to 0.3 and the hyper-parameter $\lambda$ is set to 10. More details are presented in the \textbf{supplementary material}.

\noindent \textbf{Evaluation Metrics.} For adversarial robustness evaluation, we adopt several widely used attack methods, \textit{i.e.,} PGD~\cite{madry2017towards}, C\&W~\cite{carlini2017towards}, and an ensemble of diverse parameter-free attacks, AA~\cite{croce2020reliable} that includes APGD~\cite{croce2020reliable}, APGD-T~\cite{croce2020reliable}, FAB~\cite{croce2020minimally}, and Square~\cite{andriushchenko2020square}. 
We set the maximum perturbation strength $\epsilon$ to 8 for all attack methods. 
Moreover, PGD attack is conducted with 10, 20 and, 50 iterations, \textit{i.e.}, PGD-10, PGD-20, and PGD-50. We report the results of the checkpoint with the best accuracy under the attack of PGD-10 as well as the results of the last checkpoint. 
All experiments in the manuscript are performed with the multi-step learning rate strategy. 
We also conduct experiments by using a cyclic learning rate strategy \cite{smith2017cyclical} which are presented in the \textbf{supplementary material}.

\noindent \textbf{Competing Methods.}
We compare the proposed method with a series of state-of-the-art FAT methods, \textit{i.e.,} Free-AT\cite{shafahi2019adversarial}, FGSM-RS~\cite{wong2020fast}, FGSM-GA\cite{andriushchenko2020understanding}, FGSM-CKPT\cite{kim2020understanding}, GAT~\cite{sriramanan2020guided} and NuAT~\cite{sriramanan2021towards}. 
We also compare with an advanced multi-step AT method, \textit{i.e.,} PGD-AT\cite{rice2020overfitting}), which is an early stopping version of the original PGD-based AT method~\cite{madry2017towards}.

\begin{table*}[t]
\centering
\caption{Comparisons of clean and robust accuracy (\%) and training time (minute) using ResNet18 on the CIFAR-100 dataset.  Number in bold indicates the best.
}
\label{table:cifar100_ResNet}
 \scalebox{0.85}{
\begin{tabular}{@{}c|ccccccc|c@{}}
\toprule
Method                          &      & Clean          & PGD-10         & PGD-20         & PGD-50         & C\&W           & AA             & Time(min)            \\ \midrule
\multirow{2}{*}{PGD-AT~\cite{rice2020overfitting}}          & Best & 57.52          & 29.6           & 28.99          & 28.87          & 28.85          & 25.48          & \multirow{2}{*}{284} \\ \cmidrule(lr){2-8}
                                 & Last & 57.5           & 29.54          & 29.00          & 28.90          & 27.6           & 25.48          &                      \\ \midrule \midrule
\multirow{2}{*}{FGSM-RS~\cite{wong2020fast}}         & Best & 49.85          & 22.47          & 22.01          & 21.82          & 20.55          & 18.29          & \multirow{2}{*}{70}  \\ \cmidrule(lr){2-8}
                                 & Last & 60.55          & 00.45          & 00.25          & 00.19          & 00.25          & 0.00           &                      \\ \midrule
\multirow{2}{*}{FGSM-CKPT~\cite{kim2020understanding}}       & Best & \textbf{60.93} & 16.58          & 15.47          & 15.19          & 16.4           & 14.17          & \multirow{2}{*}{96}  \\ \cmidrule(lr){2-8}
                                 & Last & \textbf{60.93} & 16.69          & 15.61          & 15.24          & 16.6           & 14.34          &                      \\ \midrule
\multirow{2}{*}{NuAT~\cite{sriramanan2020guided}}            & Best & 59.71          & 27.54          & 23.02          & 20.18          & 22.07          & 11.32          & \multirow{2}{*}{115} \\ \cmidrule(lr){2-8}
                                 & Last & 59.62          & 27.07          & 22.72          & 20.09          & 21.59          & 11.55          &                      \\ \midrule
\multirow{2}{*}{GAT~\cite{sriramanan2021towards}}             & Best & 57.01          & 24.55          & 23.8           & 23.55          & 22.02          & 19.60          & \multirow{2}{*}{119} \\ \cmidrule(lr){2-8}
                                 & Last & 56.07          & 23.92          & 23.18          & 23.0           & 21.93          & 19.51          &                      \\ \midrule
\multirow{2}{*}{FGSM-GA~\cite{andriushchenko2020understanding}}         & Best & 54.35          & 22.93          & 22.36          & 22.2           & 21.2           & 18.88          & \multirow{2}{*}{187} \\ \cmidrule(lr){2-8}
                                 & Last & 55.1           & 20.04          & 19.13          & 18.84          & 18.96          & 16.45          &                      \\ \midrule
\multirow{2}{*}{Free-AT(m=8)~\cite{shafahi2019adversarial}}    & Best & 52.49          & 24.07          & 23.52          & 23.36          & 21.66          & 19.47          & \multirow{2}{*}{229} \\ \cmidrule(lr){2-8}
                                 & Last & 52.63          & 22.86          & 22.32          & 22.16          & 20.68          & 18.57          &                      \\ \midrule
\multirow{2}{*}{FGSM-BP (ours)} & Best & 57.58 & 30.78  & 30.01  & 28.99  & 26.40 & 23.63 & \multirow{2}{*}{83}  \\ \cmidrule(lr){2-8}
                                & Last & 83.82 & 30.56  & 29.96  & 28.82  & 26.32 & 23.43 &                       \\ \bottomrule
\multirow{2}{*}{FGSM-EP (ours)} & Best & 57.74 & 31.01  & 30.17  & 29.93  & 27.37 & 24.39 & \multirow{2}{*}{83}  \\ \cmidrule(lr){2-8}
                                & Last & 57.74 & 31.01  & 30.17  & 29.93  & 27.37 & 24.39 &                      \\ \bottomrule          
\multirow{2}{*}{FGSM-MEP (ours)} & Best & 58.78          & \textbf{31.88} & \textbf{31.26} & \textbf{31.14} & \textbf{28.06} & \textbf{25.67} & \multirow{2}{*}{83}  \\ \cmidrule(lr){2-8}
                                 & Last & 58.81          & \textbf{31.6}  & \textbf{31.03} & \textbf{30.88} & \textbf{27.72} & \textbf{25.42} &                      \\ \bottomrule
\end{tabular}
}

\end{table*}
\begin{table}[t]
\centering
\caption{
Comparisons of clean and robust accuracy (\%) and training time (minute) using PreActResNet18 on Tiny ImageNet. Number in bold indicates the best.
}
\label{table:Tiny_Imagnet}
\scalebox{0.85}{
\begin{tabular}{@{}c|ccccccc|c@{}}
\toprule
Method                         &      & Clean          & PGD-10         & PGD-20        & PGD-50         & C\&W           & AA             & Time(min)             \\ \midrule
\multirow{2}{*}{PGD-AT~\cite{rice2020overfitting}}          & Best & 43.6           & 20.2           & 19.9          & 19.86          & 17.5           & 16.00          & \multirow{2}{*}{1833} \\ \cmidrule(lr){2-8}
                                 & Last & 45.28          & 16.12          & 15.6          & 15.4           & 14.28          & 12.84          &                       \\ \midrule \midrule
\multirow{2}{*}{FGSM-RS~\cite{wong2020fast}}         & Best & 44.98          & 17.72          & 17.46         & 17.36          & 15.84          & 14.08          & \multirow{2}{*}{339}  \\ \cmidrule(lr){2-8}
                                 & Last & 45.18          & 0.00           & 0.00          & 0.00           & 0.00           & 0.00           &                       \\ \midrule
\multirow{2}{*}{FGSM-CKPT~\cite{kim2020understanding}}       & Best & \textbf{49.98} & 9.20           & 9.20          & 8.68           & 9.24           & 8.10           & \multirow{2}{*}{464}  \\ \cmidrule(lr){2-8}
                                 & Last & \textbf{49.98} & 9.20           & 9.20          & 8.68           & 9.24           & 8.10           &                       \\ \midrule
\multirow{2}{*}{NuAT~\cite{sriramanan2021towards}}            & Best & 42.9           & 15.12          & 14.6          & 14.44          & 12.02          & 10.28          & \multirow{2}{*}{660}  \\ \cmidrule(lr){2-8}
                                 & Last & 42.42          & 13.78          & 13.34         & 13.2           & 11.32          & 9.56           &                       \\ \midrule
\multirow{2}{*}{GAT~\cite{sriramanan2020guided}}             & Best & 42.16          & 15.02          & 14.5          & 14.44          & 11.78          & 10.26          & \multirow{2}{*}{663}  \\ \cmidrule(lr){2-8}
                                 & Last & 41.84          & 14.44          & 13.98         & 13.8           & 11.48          & 9.74           &                       \\ \midrule
\multirow{2}{*}{FGSM-GA~\cite{andriushchenko2020understanding} }        & Best & 43.44          & 18.86           & 18.44         & 18.36            & 16.2           & 14.28           & \multirow{2}{*}{1054} \\ \cmidrule(lr){2-8}
                                 & Last& 43.44          & 18.86           & 18.44         & 18.36            & 16.2           & 14.28            &                       \\ \midrule
\multirow{2}{*}{Free-AT(m=8)~\cite{shafahi2019adversarial}}    & Best & 38.9           & 11.62          & 11.24         & 11.02          & 11.00          & 9.28           & \multirow{2}{*}{1375} \\ \cmidrule(lr){2-8}
                                 & Last & 40.06          & 8.84           & 8.32          & 8.2            & 8.08           & 7.34           &                       \\ \midrule
                                 
\multirow{2}{*}{FGSM-BP (ours)} & Best & 45.01 & 21.67  & 21.47  & 21.43  & 17.89 & 15.36 & \multirow{2}{*}{458}  \\ \cmidrule(lr){2-8}
                                 & Last & 47.16 & 20.62  & 20.16  & 20.07  & 15.68 & 14.15 &                      \\ \bottomrule  
\multirow{2}{*}{FGSM-EP (ours)} & Best & 45.01 & 21.67  & 21.47  & 21.43  & 17.89 & 15.36 & \multirow{2}{*}{458}  \\ \cmidrule(lr){2-8}
                                 & Last & 46.00 & 20.77  & 20.39  & 20.28  & 16.65 & 14.93 &                      \\ \bottomrule
\multirow{2}{*}{FGSM-MEP (ours)} & Best & 43.32          & \textbf{23.8}  & \textbf{23.4} & \textbf{23.38} & \textbf{19.28} & \textbf{17.56} & \multirow{2}{*}{458}  \\ \cmidrule(lr){2-8}
                                 & Last & 45.88          & \textbf{22.02} & \textbf{21.7} & \textbf{21.6}  & \textbf{17.44} & \textbf{15.50} &                       \\ \bottomrule
\end{tabular}
}

\end{table}
\begin{table}[]

\centering
\caption{Comparisons of clean and robust accuracy (\%) and training time (minute)  using ResNet50 on the  ImageNet dataset. Number in bold indicates the best.
}

\label{table:Imagnet}
\scalebox{0.85}{

\begin{tabular}{c|c|c|c|c|c}
\toprule 
ImageNet                          & Epsilon      & Clean          & PGD-10         & PGD-50         & Time (hour)             \\\midrule \midrule
\multirow{3}{*}{Free-AT(m=4)\cite{shafahi2019adversarial}}     & $\epsilon=$2 & 68.37          & 48.31          & 48.28          & \multirow{3}{*}{127.7} \\ 
                                  & $\epsilon=$4 & 63.42          & 33.22          & 33.08          &                        \\ 
                                  & $\epsilon=$8 & 52.09          & 19.46          & 12.92          &                        \\ \midrule
\multirow{3}{*}{FGSM-RS~\cite{wong2020fast}}          & $\epsilon=$2 & 67.65          & 48.78          & 48.67          & \multirow{3}{*}{44.5} \\ 
                                  & $\epsilon=$4 & 63.65          & 35.01          & 32.66          &                        \\ 
                                  & $\epsilon=$8 & 53.89          & 0.00           & 0.00           &                        \\ \midrule
\multirow{3}{*}{FGSM-BP (ours)} & $\epsilon=$2 & \textbf{68.41} & \textbf{49.11} & \textbf{49.10} & \multirow{3}{*}{63.7} \\ 
                                  & $\epsilon=$4 & \textbf{64.32} & \textbf{36.24} & \textbf{34.93} &                        \\ 
                                  & $\epsilon=$8 & \textbf{53.96} & \textbf{21.76} & \textbf{14.33} &                        \\ \bottomrule
\end{tabular}
}
\end{table}
\begin{table}[t]
\centering

\caption{\footnotesize Ablation study of the proposed method.}
 \label{tb:ablation}
\scalebox{0.85}{
\begin{tabular}{@{}c|c|c|c|c|c|c@{}}
\toprule
CIFAR-10                                   &      & Clean          & PGD-50         & C\&W           & AA             & Time(min)           \\ \midrule
\multirow{2}{*}{FGSM-RS}                   & Best & 73.81          & 41.26          & 39.84          & 37.07          & \multirow{2}{*}{51} \\ \cmidrule(lr){2-6}
                                           & Last & 83.82          & 00.02          & 0.00           & 0.00           &                     \\ \midrule \midrule
\multirow{2}{*}{FGSM-BP w/o regularizer}   & Best & \textbf{86.51} & 45.77          & 44.8           & 43.30          & \multirow{2}{*}{51} \\ \cmidrule(lr){2-6}
                                           & Last & 86.57 & 44.39          & 43.82          & 42.08          &                     \\ \midrule
\multirow{2}{*}{FGSM-EP w/o regularizer}   & Best & 85.97          & 45.97          & 44.6           & 43.39          & \multirow{2}{*}{51} \\ \cmidrule(lr){2-6}
                                           & Last & 86.3           & 44.97          & 43.8           & 42.84          &                     \\ \midrule
\multirow{2}{*}{FGSM-MEP w/o regularizer}  & Best & 86.33          & 46.71          & 45.5           & 43.99          & \multirow{2}{*}{51} \\ \cmidrule(lr){2-6}
                                           & Last & \textbf{86.61}          & 45.69          & 44.8           & 43.26          &                     \\ \midrule
\multirow{2}{*}{FGSM-RS with regularizer}  & Best & 84.41          & 50.63\        & 48.76          & 46.80          & \multirow{2}{*}{73} \\ \cmidrule(lr){2-6}
                                           & Last & 84.41          & 50.63          & 48.76          & 46.80          &                     \\ \midrule
\multirow{2}{*}{FGSM-BP with regularizer}  & Best & 83.15          & 53.2           & 50.24          & 47.47          & \multirow{2}{*}{73} \\ \cmidrule(lr){2-6}
                                           & Last & 83.09          & 53.33          & 50.12          & 47.17          &                     \\ \midrule
\multirow{2}{*}{FGSM-EP with regularizer}  & Best & 82.75          & 53.27          & 49.86          & 47.94          & \multirow{2}{*}{73} \\ \cmidrule(lr){2-6}
                                           & Last & 81.27          & 53.63          & 50.12          & 46.83          &                     \\ \midrule
\multirow{2}{*}{FGSM-MEP with regularizer} & Best & 81.72          & \textbf{54.17} & \textbf{50.75} & \textbf{49.00} & \multirow{2}{*}{73} \\ \cmidrule(lr){2-6}
                                           & Last & 81.72          & \textbf{54.17} & \textbf{50.75} & \textbf{49.00} &                     \\ \bottomrule
\end{tabular}
}
\end{table}

\subsection{Results on CIFAR-10} 
On CIFAR-10, we adopt ResNet18 as the backbone.  The results are shown in Table~\ref{table:cifar10_ResNet}.
Note that the results of using WideResNet34-10 as the backbone are shown in the \textbf{supplementary material}.
The proposed FGSM-MEP prevents catastrophic overfitting and 
even achieves better robustness than PGD-AT in most cases.
It costs much less time than PGD-AT. Specifically, under the AA attack, PGD-AT achieves an accuracy of about 48\% while the proposed FGSM-MEP achieves an accuracy of about 49\%. 

Compared with other FAT methods, the proposed FGSM-MEP achieves comparable robustness to the previous most potent FAT method, NuAT.
But FGSM-MEP costs less time than NuAT (73 min VS 104 min). 
Besides, for the last checkpoint, the proposed FGSM-MEP can achieve the best robustness performance under all attack scenarios. 
And for the best checkpoint, the proposed FGSM-MEP also achieves the best robustness performance under the PGD-10, PGD-20, and PGD-50.
In terms of efficiency, our training process is about 3 times faster than Free-AT, 2.5 times faster than FGSM-GA, and 1.4 times faster than GAT and NuAT. Previous fast adversarial training variants~\cite{andriushchenko2020understanding,kim2020understanding,sriramanan2020guided,sriramanan2021towards} improve the quality of adversarial examples in different ways. Though they can prevent catastrophic overfitting and improve model robustness, they all require much time for  quality improvement. 
Differently, we improve adversarial example quality from the perspective of initialization and propose to adopt historically generated adversarial perturbation to initialize the adversarial examples without additional calculation cost.


\subsection{Results on CIFAR-100}  
On CIFAR-100, we adopt ResNet18 as the backbone. 
The results are shown in Table~\ref{table:cifar100_ResNet}.
Compared with CIFAR-10, it is hard for the classification model to obtain robustness because the CIFAR-100 covers more classes. 
The proposed FGSM-MEP achieves comparable robustness to PGD-AT and costs less time than PGD-AT. Specifically, under the AA attack,  PGD-AT achieves an
accuracy of about 25\%, while our FGSM-MEP also achieves an
accuracy of about 25\%. 
Note that our training process is about 3 times faster than PGD-AT.
Compared with other FAT methods, 
the proposed FGSM-MEP also achieves the best robustness against AEs under all attack scenarios on the best and last checkpoints.
In terms of training efficiency, we observe similar results on CIFAR-10.

\subsection{Results on Tiny ImageNet and ImageNet} 

\noindent \textbf{Results on Tiny ImageNet.}  Following the setting of \cite{kim2020understanding,rice2020overfitting}, we adopt PreActResNet18 to conduct AT. 
The results are shown in Table~\ref{table:Tiny_Imagnet}. 
Compared with competing FAT methods, our FGSM-MEP achieves higher robust accuracy.
Compared with PGD-AT, our FGSM-MEP achieves better robustness performance under all attack scenarios on the best and last checkpoints. In terms of training efficiency, we observe similar results on CIFAR-10 and CIFAR-100.

\noindent \textbf{Results on ImageNet.} 
Following the setting of \cite{shafahi2019adversarial,wong2020fast}, we adopt ResNet50 to conduct experiments. 
Specifically, ResNet50 is trained with the maximum perturbation strength $\epsilon=2$, $\epsilon=4$, and $\epsilon=8$. The proposed FGSM-EP and FGSM-MEP require memory consumption to store the adversarial perturbation of the last epoch, which limits their application on ImageNet. Fortunately, FGSM-BP does not require memory consumption to conduct AT. Hence, on the ImageNet,
we compare our FGSM-BP with FGSM-RS and Free-AT. The results are shown in Table~\ref{table:Imagnet}. 
Compared with FGSM-RS and Free-AT, our FGSM-BP achieves the highest clean and robust accuracy. 
Our FGSM-BP requires a bit more calculation cost than FGSM-RS, but much less time than Free-AT.

\subsection{Ablation Study}
In this paper, we propose a regularization loss term to enforce function smoothness, resulting in improving model robustness. 
To validate the effectiveness of the proposed regularization, we adopt ResNet18 as the classification model to conduct ablation experiments on CIFAR-10. 
The results are shown in Table~\ref{tb:ablation}. It can be observed that combined with our regularization method, FGSM-BP, FGSM-EP, and FGSM-MEP can achieve better robustness performance under all attack scenarios. 


\section{Conclusion}
In this paper, we investigate how to improve adversarial example quality from the perspective of initialization and propose to adopt historically generated adversarial perturbations to initialize adversarial examples. 
It can generate powerful adversarial examples with no additional calculation cost. 
Moreover, we also propose a simple yet effective regularizer to further improve model robustness, which prevents the current perturbation deviating too much from the prior-guided initialization.
The regularizer adopts both historical and current adversarial perturbations to guide the model learning. Extensive experimental evaluations demonstrate that the proposed method can prevent catastrophic overfitting and outperform state-of-the-art FAT methods at a low computational cost.




~\\
\noindent \textbf{Acknowledgement}\\
Supported by the National Key R\&D Program of China under Grant 2018AAA01\\02503, National Natural Science Foundation of China (No. U2001202, U1936208, 62006217). Beijing Natural Science Foundation (No. M22006).  Shenzhen Science and Technology Program under grant No.RCYX20210609103057050, and Tencent AI Lab Rhino-Bird Focused Research Program under grant No. JR202123.
\clearpage
%
%
\bibliographystyle{splncs04}
\bibliography{egbib}

\begin{thebibliography}{10}
\providecommand{\url}[1]{\texttt{#1}}
\providecommand{\urlprefix}{URL }
\providecommand{\doi}[1]{https://doi.org/#1}

\bibitem{andriushchenko2020square}
Andriushchenko, M., Croce, F., Flammarion, N., Hein, M.: Square attack: a
  query-efficient black-box adversarial attack via random search. In: European
  Conference on Computer Vision. pp. 484--501. Springer (2020)

\bibitem{andriushchenko2020understanding}
Andriushchenko, M., Flammarion, N.: Understanding and improving fast
  adversarial training. Advances in Neural Information Processing Systems
  \textbf{33},  16048--16059 (2020)

\bibitem{bai2020targeted}
Bai, J., Chen, B., Li, Y., Wu, D., Guo, W., Xia, S.t., Yang, E.h.: Targeted
  attack for deep hashing based retrieval. In: European Conference on Computer
  Vision. pp. 618--634. Springer (2020)

\bibitem{bai2020improving}
Bai, Y., Zeng, Y., Jiang, Y., Wang, Y., Xia, S.T., Guo, W.: Improving query
  efficiency of black-box adversarial attack. In: European Conference on
  Computer Vision. pp. 101--116. Springer (2020)

\bibitem{bai2021improving}
Bai, Y., Zeng, Y., Jiang, Y., Xia, S.T., Ma, X., Wang, Y.: Improving
  adversarial robustness via channel-wise activation suppressing. arXiv
  preprint arXiv:2103.08307  (2021)

\bibitem{DBLP:conf/iclr/BaiZJXM021}
Bai, Y., Zeng, Y., Jiang, Y., Xia, S., Ma, X., Wang, Y.: Improving adversarial
  robustness via channel-wise activation suppressing. In: 9th International
  Conference on Learning Representations, {ICLR} 2021, Virtual Event, Austria,
  May 3-7, 2021. OpenReview.net (2021)

\bibitem{carlini2017towards}
Carlini, N., Wagner, D.: Towards evaluating the robustness of neural networks.
  In: 2017 ieee symposium on security and privacy (sp). pp. 39--57. Ieee (2017)

\bibitem{chen2018robust}
Chen, S.T., Cornelius, C., Martin, J., Chau, D.H.: Robust physical adversarial
  attack on faster r-cnn object detector. corr abs/1804.05810 (2018). arXiv
  preprint arXiv:1804.05810  (2018)

\bibitem{croce2020minimally}
Croce, F., Hein, M.: Minimally distorted adversarial examples with a fast
  adaptive boundary attack. In: International Conference on Machine Learning.
  pp. 2196--2205. PMLR (2020)

\bibitem{croce2020reliable}
Croce, F., Hein, M.: Reliable evaluation of adversarial robustness with an
  ensemble of diverse parameter-free attacks. In: International conference on
  machine learning. pp. 2206--2216. PMLR (2020)

\bibitem{deng2009imagenet}
Deng, J., Dong, W., Socher, R., Li, L.J., Li, K., Fei-Fei, L.: Imagenet: A
  large-scale hierarchical image database. In: 2009 IEEE conference on computer
  vision and pattern recognition. pp. 248--255. Ieee (2009)

\bibitem{duan2020adversarial}
Duan, R., Ma, X., Wang, Y., Bailey, J., Qin, A.K., Yang, Y.: Adversarial
  camouflage: Hiding physical-world attacks with natural styles. In:
  Proceedings of the IEEE/CVF conference on computer vision and pattern
  recognition. pp. 1000--1008 (2020)

\bibitem{duan2021adversarial}
Duan, R., Mao, X., Qin, A.K., Chen, Y., Ye, S., He, Y., Yang, Y.: Adversarial
  laser beam: Effective physical-world attack to dnns in a blink. In:
  Proceedings of the IEEE/CVF Conference on Computer Vision and Pattern
  Recognition. pp. 16062--16071 (2021)

\bibitem{finlayson2019adversarial}
Finlayson, S.G., Bowers, J.D., Ito, J., Zittrain, J.L., Beam, A.L., Kohane,
  I.S.: Adversarial attacks on medical machine learning. Science
  \textbf{363}(6433),  1287--1289 (2019)

\bibitem{goodfellow2014explaining}
Goodfellow, I.J., Shlens, J., Szegedy, C.: Explaining and harnessing
  adversarial examples (2014)

\bibitem{gu2021capsule}
Gu, J., Tresp, V., Hu, H.: Capsule network is not more robust than
  convolutional network. In: Proceedings of the IEEE/CVF Conference on Computer
  Vision and Pattern Recognition. pp. 14309--14317 (2021)

\bibitem{gu2021effective}
Gu, J., Wu, B., Tresp, V.: Effective and efficient vote attack on capsule
  networks. arXiv preprint arXiv:2102.10055  (2021)

\bibitem{he2016deep}
He, K., Zhang, X., Ren, S., Sun, J.: Deep residual learning for image
  recognition. In: Proceedings of the IEEE conference on computer vision and
  pattern recognition. pp. 770--778 (2016)

\bibitem{he2016identity}
He, K., Zhang, X., Ren, S., Sun, J.: Identity mappings in deep residual
  networks. In: European conference on computer vision. pp. 630--645. Springer
  (2016)

\bibitem{jia2019comdefend}
Jia, X., Wei, X., Cao, X., Foroosh, H.: Comdefend: An efficient image
  compression model to defend adversarial examples. In: Proceedings of the
  IEEE/CVF conference on computer vision and pattern recognition. pp.
  6084--6092 (2019)

\bibitem{jia2020adv}
Jia, X., Wei, X., Cao, X., Han, X.: Adv-watermark: A novel watermark
  perturbation for adversarial examples. In: Proceedings of the 28th ACM
  International Conference on Multimedia. pp. 1579--1587 (2020)

\bibitem{jia2022adversarial}
Jia, X., Zhang, Y., Wu, B., Ma, K., Wang, J., Cao, X.: Las-at: Adversarial
  training with learnable attack strategy. In: Proceedings of the IEEE/CVF
  Conference on Computer Vision and Pattern Recognition. pp. 13398--13408
  (2022)

\bibitem{jia2022boosting}
Jia, X., Zhang, Y., Wu, B., Wang, J., Cao, X.: Boosting fast adversarial
  training with learnable adversarial initialization. IEEE Transactions on
  Image Processing  (2022)

\bibitem{kannan2018adversarial}
Kannan, H., Kurakin, A., Goodfellow, I.: Adversarial logit pairing. arXiv
  preprint arXiv:1803.06373  (2018)

\bibitem{kim2020understanding}
Kim, H., Lee, W., Lee, J.: Understanding catastrophic overfitting in
  single-step adversarial training. In: Proceedings of the AAAI Conference on
  Artificial Intelligence. vol.~35, pp. 8119--8127 (2021)

\bibitem{krizhevsky2009learning}
Krizhevsky, A., Hinton, G., et~al.: Learning multiple layers of features from
  tiny images  (2009)

\bibitem{lecun2015deep}
LeCun, Y., Bengio, Y., Hinton, G.: Deep learning. nature  \textbf{521}(7553),
  436--444 (2015)

\bibitem{li2022semi}
Li, Y., Wu, B., Feng, Y., Fan, Y., Jiang, Y., Li, Z., Xia, S.T.:
  Semi-supervised robust training with generalized perturbed neighborhood.
  Pattern Recognition  \textbf{124},  108472 (2022)

\bibitem{liang2020efficient}
Liang, S., Wei, X., Yao, S., Cao, X.: Efficient adversarial attacks for visual
  object tracking. In: European Conference on Computer Vision. pp. 34--50.
  Springer (2020)

\bibitem{liang2022parallel}
Liang, S., Wu, B., Fan, Y., Wei, X., Cao, X.: Parallel rectangle flip attack: A
  query-based black-box attack against object detection. arXiv preprint
  arXiv:2201.08970  (2022)

\bibitem{lin2019nesterov}
Lin, J., Song, C., He, K., Wang, L., Hopcroft, J.E.: Nesterov accelerated
  gradient and scale invariance for adversarial attacks. arXiv preprint
  arXiv:1908.06281  (2019)

\bibitem{madry2017towards}
Madry, A., Makelov, A., Schmidt, L., Tsipras, D., Vladu, A.: Towards deep
  learning models resistant to adversarial attacks. arXiv preprint
  arXiv:1706.06083  (2017)

\bibitem{pang2020bag}
Pang, T., Yang, X., Dong, Y., Su, H., Zhu, J.: Bag of tricks for adversarial
  training. In: 9th International Conference on Learning Representations,
  {ICLR} 2021, Virtual Event, Austria, May 3-7, 2021. OpenReview.net (2021)

\bibitem{pang2020boosting}
Pang, T., Yang, X., Dong, Y., Xu, K., Zhu, J., Su, H.: Boosting adversarial
  training with hypersphere embedding. arXiv preprint arXiv:2002.08619  (2020)

\bibitem{park2021reliably}
Park, G.Y., Lee, S.W.: Reliably fast adversarial training via latent
  adversarial perturbation. arXiv preprint arXiv:2104.01575  (2021)

\bibitem{DBLP:journals/nn/Qian99}
Qian, N.: On the momentum term in gradient descent learning algorithms. Neural
  networks  \textbf{12}(1),  145--151 (1999)

\bibitem{rice2020overfitting}
Rice, L., Wong, E., Kolter, Z.: Overfitting in adversarially robust deep
  learning. In: International Conference on Machine Learning. pp. 8093--8104.
  PMLR (2020)

\bibitem{roth2019adversarial}
Roth, K., Kilcher, Y., Hofmann, T.: Adversarial training is a form of
  data-dependent operator norm regularization. arXiv preprint arXiv:1906.01527
  (2019)

\bibitem{shafahi2019adversarial}
Shafahi, A., Najibi, M., Ghiasi, M.A., Xu, Z., Dickerson, J., Studer, C.,
  Davis, L.S., Taylor, G., Goldstein, T.: Adversarial training for free!
  Advances in Neural Information Processing Systems  \textbf{32} (2019)

\bibitem{smith2017cyclical}
Smith, L.N.: Cyclical learning rates for training neural networks. In: 2017
  IEEE winter conference on applications of computer vision (WACV). pp.
  464--472. IEEE (2017)

\bibitem{sriramanan2020guided}
Sriramanan, G., Addepalli, S., Baburaj, A., et~al.: Guided adversarial attack
  for evaluating and enhancing adversarial defenses. Advances in Neural
  Information Processing Systems  \textbf{33},  20297--20308 (2020)

\bibitem{sriramanan2021towards}
Sriramanan, G., Addepalli, S., Baburaj, A., et~al.: Towards efficient and
  effective adversarial training. Advances in Neural Information Processing
  Systems  \textbf{34} (2021)

\bibitem{szegedy2013intriguing}
Szegedy, C., Zaremba, W., Sutskever, I., Bruna, J., Erhan, D., Goodfellow, I.,
  Fergus, R.: Intriguing properties of neural networks. arXiv preprint
  arXiv:1312.6199  (2013)

\bibitem{wang2021enhancing}
Wang, X., He, K.: Enhancing the transferability of adversarial attacks through
  variance tuning. In: Proceedings of the IEEE/CVF Conference on Computer
  Vision and Pattern Recognition. pp. 1924--1933 (2021)

\bibitem{DBLP:conf/icml/WangM0YZG19}
Wang, Y., Ma, X., Bailey, J., Yi, J., Zhou, B., Gu, Q.: On the convergence and
  robustness of adversarial training. In: Chaudhuri, K., Salakhutdinov, R.
  (eds.) Proceedings of the 36th International Conference on Machine Learning,
  {ICML} 2019, 9-15 June 2019, Long Beach, California, {USA}. pp. 6586--6595.
  {PMLR} (2019)

\bibitem{wang2019improving}
Wang, Y., Zou, D., Yi, J., Bailey, J., Ma, X., Gu, Q.: Improving adversarial
  robustness requires revisiting misclassified examples. In: International
  Conference on Learning Representations (2019)

\bibitem{DBLP:conf/iclr/0001ZY0MG20}
Wang, Y., Zou, D., Yi, J., Bailey, J., Ma, X., Gu, Q.: Improving adversarial
  robustness requires revisiting misclassified examples. In: 8th International
  Conference on Learning Representations, {ICLR} 2020, Addis Ababa, Ethiopia,
  April 26-30, 2020. OpenReview.net (2020)

\bibitem{wei2018transferable}
Wei, X., Liang, S., Chen, N., Cao, X.: Transferable adversarial attacks for
  image and video object detection. arXiv preprint arXiv:1811.12641  (2018)

\bibitem{wong2020fast}
Wong, E., Rice, L., Kolter, J.Z.: Fast is better than free: Revisiting
  adversarial training. arXiv preprint arXiv:2001.03994  (2020)

\bibitem{DBLP:conf/nips/WuX020}
Wu, D., Xia, S.T., Wang, Y.: Adversarial weight perturbation helps robust
  generalization. vol.~33, pp. 2958--2969 (2020)

\bibitem{xie2019improving}
Xie, C., Zhang, Z., Zhou, Y., Bai, S., Wang, J., Ren, Z., Yuille, A.L.:
  Improving transferability of adversarial examples with input diversity. In:
  Proceedings of the IEEE/CVF Conference on Computer Vision and Pattern
  Recognition. pp. 2730--2739 (2019)

\bibitem{DBLP:conf/bmvc/ZagoruykoK16}
Zagoruyko, S., Komodakis, N.: Wide residual networks (2016)

\bibitem{zhang2019you}
Zhang, D., Zhang, T., Lu, Y., Zhu, Z., Dong, B.: You only propagate once:
  Accelerating adversarial training via maximal principle. arXiv preprint
  arXiv:1905.00877  (2019)

\bibitem{zhang2019theoretically}
Zhang, H., Yu, Y., Jiao, J., Xing, E., El~Ghaoui, L., Jordan, M.: Theoretically
  principled trade-off between robustness and accuracy. In: International
  Conference on Machine Learning. pp. 7472--7482. PMLR (2019)

\bibitem{zou2019reinforced}
Zou, W., Huang, S., Xie, J., Dai, X., Chen, J.: A reinforced generation of
  adversarial examples for neural machine translation. arXiv preprint
  arXiv:1911.03677  (2019)

\end{thebibliography}
\end{document}